\documentclass[conference]{IEEEtran}

\usepackage{url}
\usepackage{cite}
\usepackage{amsmath,amssymb,amsfonts}
\usepackage{bbm}
\usepackage{algorithmic}
\usepackage{graphicx}
\usepackage{textcomp}
\usepackage{booktabs}
\usepackage[table]{xcolor}

\definecolor{lightPurple}{rgb}{0.90,1,0.90}
\def\BibTeX{{\rm B\kern-.05em{\sc i\kern-.025em b}\kern-.08em
    T\kern-.1667em\lower.7ex\hbox{E}\kern-.125emX}}
\begin{document}

\title{TraceNet: Segment One Thing Efficiently}

\author{
\IEEEauthorblockN{Mingyuan Wu\textsuperscript{1}, Zichuan Liu\textsuperscript{2}, Haozhen Zheng\textsuperscript{1}, Hongpeng Guo\textsuperscript{1}, Bo Chen\textsuperscript{1}, Xin Lu\textsuperscript{2}, Klara Nahrstedt\textsuperscript{1}}
\IEEEauthorblockA{\textit{\textsuperscript{1}Coordinated Science Laboratory, University of Illinois at Urbana-Champaign, Champaign, USA, \textsuperscript{1}Independent Researcher, USA}}
\textit{\{mw34,haozhen3,hg5,boc2,klara\}@illinois.edu, \{sun8878232,xinlu.psu\}@gmail.com}\vspace{-7.5mm}}

\maketitle

\begin{abstract}
Efficient single instance segmentation is critical for unlocking features in on-the-fly mobile imaging applications, such as photo capture and editing. Existing mobile solutions often restrict segmentation to portraits or salient objects due to computational constraints. Recent advancements like the Segment Anything Model improve accuracy but remain computationally expensive for mobile, because it processes the entire image with heavy transformer backbones. To address this, we propose TraceNet, a one-click-driven single instance segmentation model. TraceNet segments a user-specified instance by back-tracing the receptive field of a ConvNet backbone, focusing computations on relevant regions and reducing inference cost and memory usage during mobile inference. Starting from user needs in real mobile applications, we define efficient single-instance segmentation tasks and introduce two novel metrics to evaluate both accuracy and robustness to low-quality input clicks. Extensive evaluations on the MS-COCO and LVIS datasets highlight TraceNet’s ability to generate high-quality instance masks efficiently and accurately while demonstrating robustness to imperfect user inputs. 
\end{abstract}


\section{Introduction}
\begin{figure}[t]
	\centering
	\includegraphics[width=0.5\textwidth]{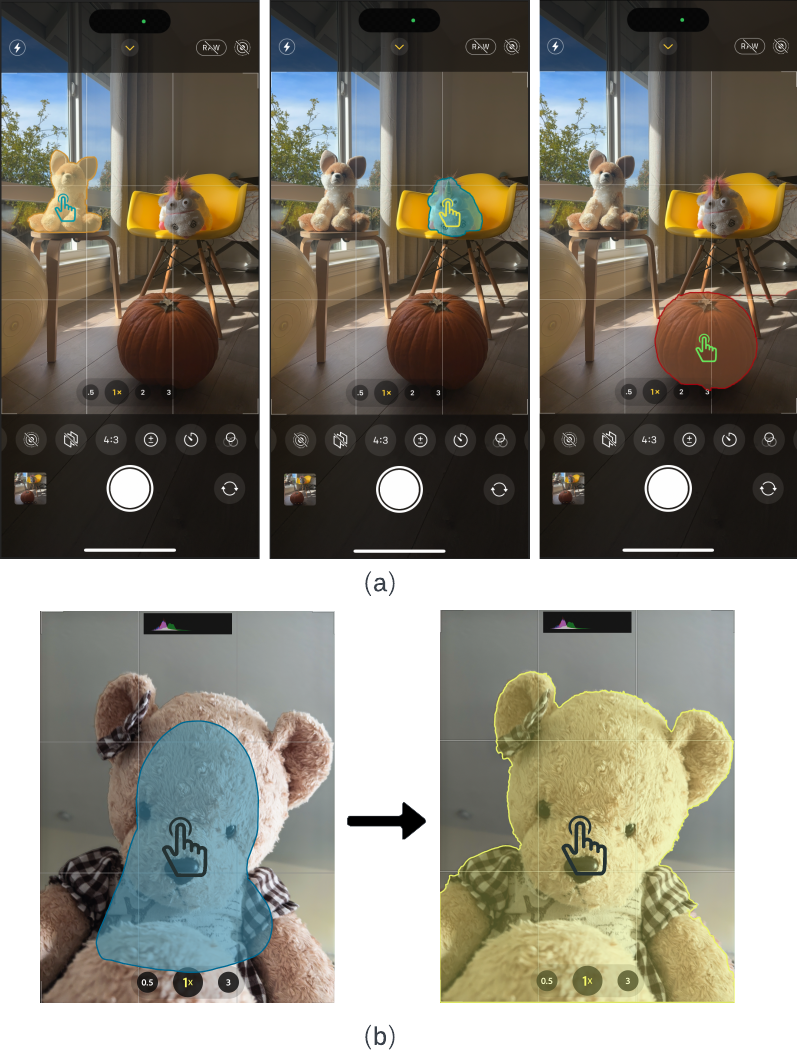}
	\caption{(a) IOS viewfinder: segment one object. (b) Blue: Clickable regions supported by our algorithm; Yellow: Predicted Mask. Note: When a user taps an object, the system (in your phone) processes this input as a pixel "click,"}
	\label{fig:intro}
\end{figure}

In recent years, the rapid development of mobile devices and applications has driven researchers to explore efficient segmentation methods that enable smooth performance in mobile editing or capturing apps. Efficient segmentation approaches \cite{sipMask,condinst,yolact-plus-tpami2020} usually build upon efficient neural network architectures \cite{ mobilenetv3}, limit the number of semantic categories, or focus on the salient subject in the scene. One successful application of efficient segmentation approaches is the portrait mode that is widely supported in default camera apps on mobile phones\footnote{iphone portrait mode: \url{https://support.apple.com/en-us/HT208118}, Pixel6 portrait mode: \url{https://store.google.com/us/magazine/pixel_camera?hl=en-US}}, which leverages the portrait segmentation technique together with the lens blur algorithms. To further enrich the set of imaging tools in the mobile capturing and editing softwares, we propose to use intuitive user inputs (such as an arbitrary click on the instance) to enable efficient single-instance segmentation. The proposed formulation relaxes the limitation of existing efficient segmentation that applies only to the salient subject in the scene. As shown in Figure \ref{fig:intro}(a), with a click of the dog, unicorn, or pumpkin, we expect to get its segmentation mask instantly. With the click-based single-instance segmentation, automatic images manipulation, such as depth-of-field effects, background replacement and image enhancement, therefore can be enabled on an arbitrary instance in the image.

Incorporating user clicks into efficient segmentation tasks poses unique challenges. User clicks are inherently random and can land anywhere within the instance's region, making it difficult to derive a precise bounding box to locate the instance. One straightforward solution is to run an efficient instance segmentation model to produce masks for all instances in the image, followed by querying the desired mask based on the user click. However, this approach wastes computational resources by generating masks for irrelevant instances. Alternative methods include click-based interactive segmentation and promptable segmentation models. While these approaches can incorporate click prompts, they often fail to guarantee efficiency. For instance, many require computationally expensive processes, such as generating features across the entire image or performing multiple optimization stages. A prominent example is Segment Anything Model \cite{sam}, which uses a transformer-based image encoder to produce a full image embedding. While powerful, this results in an inference time of approximately 0.15 seconds on an Nvidia A100 GPU, making it impractical for mobile devices due to the significant computational overhead.

In this paper, we step back from the most popular SAM research that targets at a more powerful and generalizable segmentation model. Instead, we study on \textbf{one click efficient segmentation} which is motivated by the demands of on-the-fly image editing mobile applications. The task aims at efficiently segmenting one instance that a user queries with a positive click. In practical scenarios of one-click efficient segmentation, a user's click might not always accurately target the center of the desired instance. To improve user experience, we propose a metric of user click tolerance in addition to segmentation accuracy. This metric measures the proportion of the region that a user tap can fall into for a high-quality single-instance mask, as shown in Figure \ref{fig:intro} (b). 

Our key insight into the proposed problem is that a user's click acts as a strong prior in the one-click segmentation task. The click explicitly indicates the presence of a single object of interest at the specified position. By leveraging this prior, the model can effectively eliminate redundancy and achieve significant efficiency gains. Building on this idea, we revisit convolutional networks (ConvNets) for efficient segmentation and refine the concept of the \textit{receptive field} to precisely locate and crop redundant regions at a fine-grained level, and thereby avoiding the need for intensive feature computation across the entire image during inference. Central to our segmentation framework is a novel component called the Receptive Field Tracer (RFT). The RFT back-traces computational dependencies across ConvNet layers, managing feature regions where heavy computations are applied and preserving only the activated neuron components in the forwarding path. 

To directly condition the final segmentation results directly on these modified feature regions, we build our framework upon recently proposed conditional instance segmentation methods \cite{condinst}. These methods predict local query features around the selected instance and dynamically condition the instance-aware mask head on these localized features. By integrating the RFT and ConvNet-based segmentation, our framework, named \textbf{TraceNet}, focuses computational resources exclusively on regions relevant to the target instance. Benefited from the RFT and the conditional ConvNet design, TraceNet achieves high accuracy, efficiency, and robustness in one instance segmentation tasks, making it highly suitable for deployment on resource-constrained mobile devices.

Overall, our contribution can be summarized as follows: 
\begin{itemize}
    \item Motivated by real-world demands, we proposed and formulated a \textbf{one click efficient segmentation} as a new form of efficient segmentation for a single instance, and designed evaluation protocols.
    \item We propose a solution of TraceNet for one click efficient segmentation. TraceNet conditions the instance-aware mask head on local features around the user's click and efficiently controls the usage of local features.
    \item We evaluated the proposed TraceNet on MS-COCO \cite{coco} and LVIS \cite{lvis}. TraceNet demonstrates high computational efficiency while achieving high accuracy on the mask prediction of user-specified instances along with a high user click tolerance.
\end{itemize}

\section{Related Work}
\noindent The proposed problem of one tap efficient segmentation aligns closely with a broad line of segmentation research that leverages user clicks.

\noindent \textbf{Click-based Interactive Segmentation.} Most of existing interactive segmentation pipelines optimize for the minimal user clicks so that the IoU (Intersection over Union) between the predicted foreground mask and the groundtruth mask exceeds a pre-defined threshold.  Classic methods in this field typically utilize low-level image features and the properties of clicks, such as Graphcut \cite{graphcut1} and Intelligent Scissors \cite{intelligent, interactive_intelligent}. In contrast, CNN-based models \cite{brs,fbrs2020,pseudoClick_arxiv,attention,focalclick,RITM,new_int,simpleclick} encode the click map and concatenate with RGB channel as inputs of neural networks.

\noindent \textbf{Promptable Segmentation.}
In addition to these specialized interactive segmentation models, vision foundation models have also demonstrated capability in this area. SAM \cite{sam}, in particular, has attracted significant attention for its remarkable zero-shot generalization to new image distributions and tasks. SAM operates by conditioning its mask decoder on combined output embeddings from a heavy image encoder and a prompt encoder, inherently functioning as an interactive segmenter for any instance. MobileSAM \cite{mobilesam} makes SAM more mobile-friendly by distilling the knowledge from SAM's heavy encoder into a more lightweight one. 

\noindent However, our task differs from conventional interactive segmentation, as it focuses on efficiently segmenting a single instance rather than segmenting anything or everything. Guided by a user's tap within one instance, TraceNet can selectively encode the instance/tap-relevant features and thus benefit from efficiency gains as it avoids redundant feature calculation across the entire image. This makes it well-suited for latency-sensitive applications—such as one-tap segmentation on mobile devices or in teleconferencing scenarios \cite{tele_1, tele_2}—where real-time performance is critical.

\section{Task and Method}
\label{sec:approach}
\begin{figure}[]
	\centering
	\includegraphics[width=\linewidth]{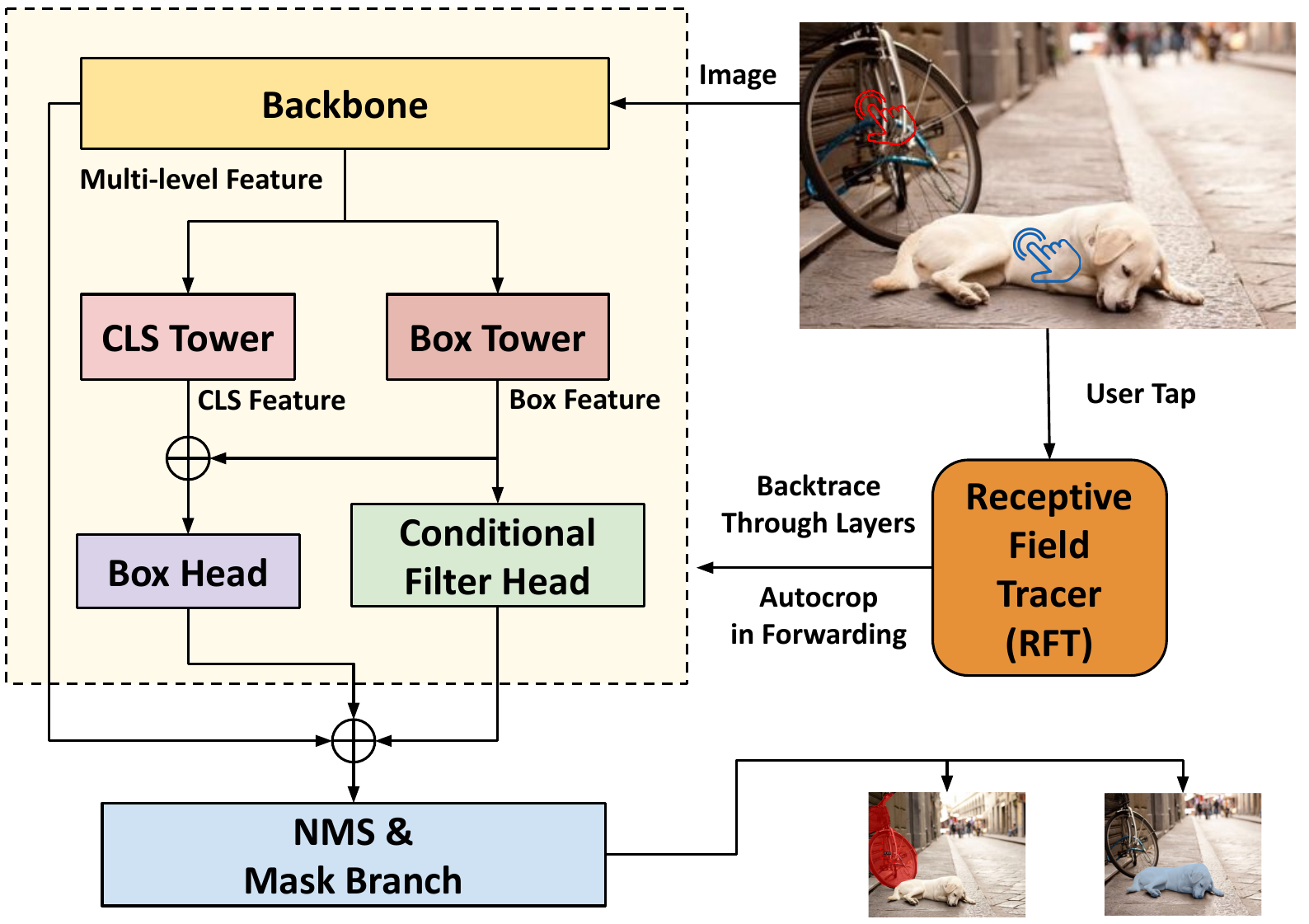}
	\caption{- Overall architecture of TraceNet. Receptive Field Tracer (RFT) back traces the receptive field region and perform autocropping in ConvNet.}
	\label{fig:arch}
\end{figure}

\noindent \textbf{Problem Formulation}. Given an input image $I \in \mathbb{R}^{H \times W \times 3}$ and a user click $c =(x, y) \in \mathbb{R}^{H \times W}$. The goal of click-driven single-instance segmentation is to predict a pixel-level mask of the instance located at $(x, y)$. The ground-truth is defined by $\left\{M_{gt}\right\}$, where $M_{gt} \in\{0,1\}^{H \times W}$ is the groundtruth mask that user queries with the click $c$, where $c \in M_{gt}$. The expected model is a learned function that maps $I$ and $c$ to a single-instance mask $M_{pred}$, targeting a satisfying IoU between $M_{pred}$ and $M_{gt}$, and robustness w.r.t the click $c$.

\noindent \textbf{TraceNet Overview}. As illustrated in Figure 2, TraceNet comprises a Receptive Field Tracer (RFT) and a ConvNet module for instance segmentation.When a user taps on the screen, the device's hardware processes the signal and converts it into the digital representation of a single pixel. The ConvNet model is pre-loaded on the device, while the RFT dynamically influences computations by backtracking the ConvNet's receptive field based on the tap location and performing automatic cropping. Further details about the RFT are provided in \ref{sec:rft}, with the ConvNet module described in \ref{sec:conv}.

\subsection{Receptive Field Tracer (RFT) in TraceNet}
\label{sec:rft}
\noindent \textbf{RFT Overview}. With the user click information, the model can identify the target instance and focuses computation accordingly, reducing unnecessary processing in other regions. To achieve this, we introduce RFT, designed to eliminate redundant computations inherent in exhaustive instance searches in existing segmentation algorithms. As shown in Figure~\ref{fig:arch}, RFT uses the user click as input, backtracks the receptive field (in blue grids) across model layers in the computational graph, and autocrops regions (in gray background) that do not contribute to the output associated with the click. This method significantly enhances memory efficiency and inference speed. Additionally, RFT is highly compatible with most ConvNet-based instance segmentation algorithms, as it does not impose restrictive architectural requirements.

\begin{figure}[]
	\centering
	\includegraphics[width=\linewidth]{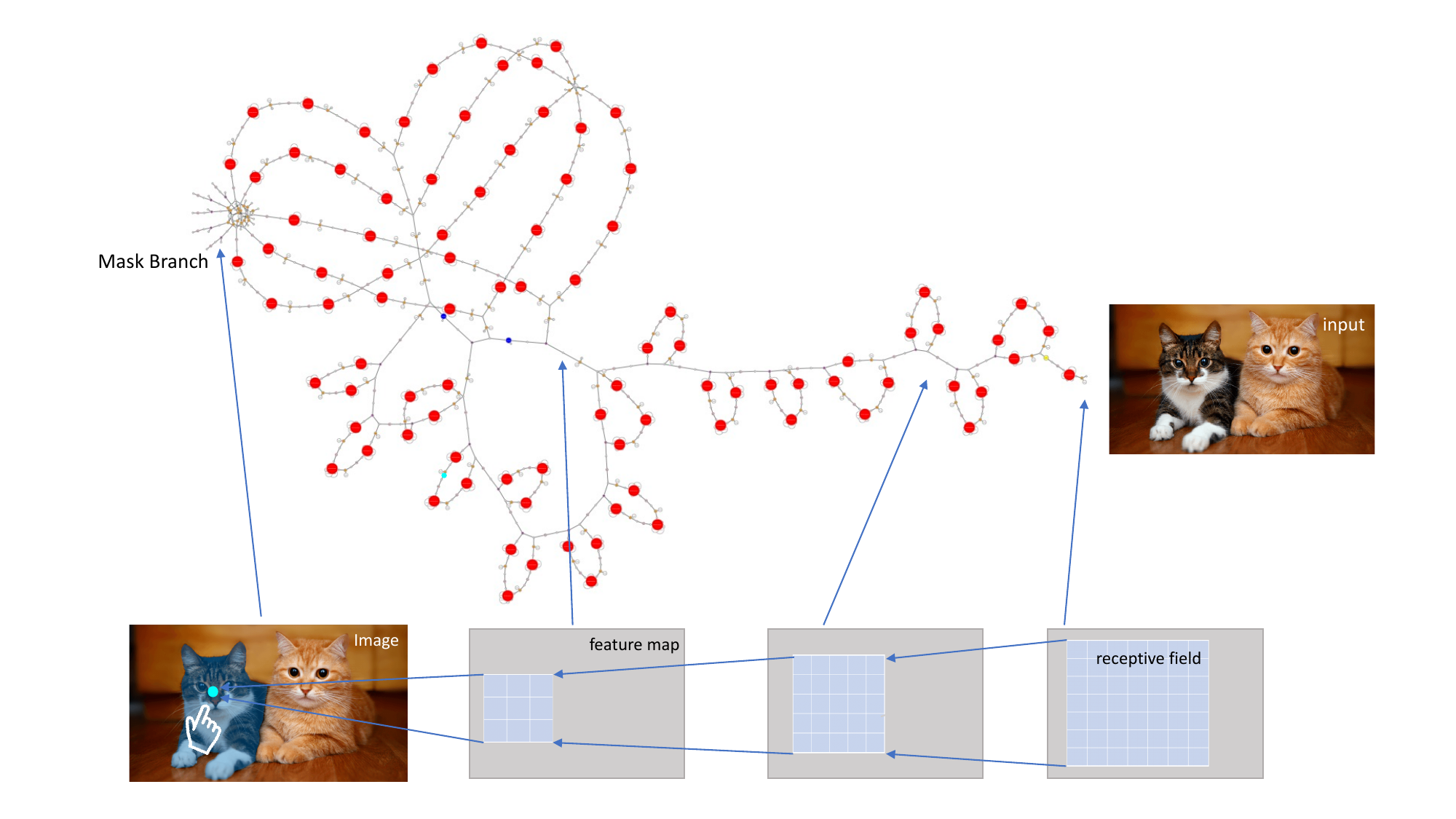}
	\caption{One click driven receptive field back tracing in the acyclic computation graph (visualized directly by onnx) of instance segmentation.}
	\label{fig:RFT}
\end{figure}

\noindent \textbf{Back-tracing Mechanism}. Designing an algorithm for receptive field back-tracing is a non-trivial task, as it must efficiently traverse layers while propagating the click information to guide computation. To address this, we propose a Depth-First Search (DFS) algorithm for backtracing in ConvNets. As illustrated in Figure~\ref{fig:RFT}, the receptive field of each layer in the computational graph for instance segmentation is recursively backtraced using DFS, starting from the downstream layers and moving toward the upstream layers.

Details will be discussed in the following sections: first, we provide a formal definition of the receptive field region at each layer of a neural network. Next, we describe the post-order DFS algorithm, illustrating how receptive field back-tracing operates in arbitrary pure convolutional neural network models, deriving all recursive cases. Finally, we explain how RFT leverages the results to manage computations in other components during inference, including autocropping and autopadding. A detailed algorithm block for RFT is provided in the Appendix.

\noindent \textbf{Receptive Field Revisited}. 
In the context of deep learning \cite{RFcompute, effectiveRF}, \emph{receptive field region} refers to the region in the input that produces the feature. \emph{Receptive field} is defined as the size of the region. The concept of receptive field is important for researchers to diagnose how CNN works in a sense that a unit in the model output is only affected by units in receptive field regions in the input image. A formal definition of the receptive field region of a simple $n$-layer convolutional neural network can be formulated as follows. Assume the pixels on each layer are indexed by $(i, j)$, with the most upper-left pixel at $(0, 0)$. Denote the $(i, j)$th pixel on the $p$-th layer as $x_{i, j}^p$. $p \in[n]$ in a $n$-layer convolutional neural network where $x_{i, j}^0$ and $x_{i, j}^n$ respectively denote the pixel value in the input image and the model output. By definition the receptive field region of the unit $x_{i, j}^n$ is the set of all units in $x^0$ that contribute to $x_{i, j}^n$. The receptive field region of a set of units is the union of the receptive field region of all units in the set. We extend the concept of the receptive field more than in the input images and define \textit{the $p$-layer receptive field region $r^p$} of the unit $x_{i, j}^n$ to be the set of all units in the output feature map of the pth layer $x^p$ that contribute to $x_{i, j}^n$, for any $p \in[n]$. Note that we can consider only single channel of the input and output of each layer in the context of calculating receptive field regions and similar results can be derived for layers with multiple channels.

\noindent \textbf{Receptive Field Depth First Search}.
The click-driven segmentation aims at precisely localizing the region that contributes to local output features around user clicks and reducing spatial redundancy of convolution operations as much as possible across all the layers of the model. i.e only computation within the layer receptive field region is preserved in forward pass. We introduce a click-driven receptive field back-tracing algorithm to compute receptive field regions at each layer of the neural network from deep to shallow. To illustrate the algorithm, we construct a directed acyclic computation graph for arbitrary model, where the nodes correspond to layer operations and inputs, and directed edges represent dependency between layers. (Note that most modern convolutional neural networks designs rely on layers with multiple child nodes and more than one back-tracing paths exist.) To deal with arbitrary neural network model, the calculation of receptive field region at each layer is conducted in post-order Depth First Search with an intuition that $p$-layer receptive field region can be represented as a function of receptive field regions of all the child nodes of the $p$-layer node. In post-order Depth First Search, the receptive field region of the $p$-layer will be calculated after all the child nodes of the $p$-layer have been visited. The search algorithm ensures single visit of each node and results in a worst-case complexity of $\mathcal{O}(|\mathcal{E}|+|\mathcal{N}|)$, where $\mathcal{E}$ represents the edge set and $\mathcal{N}$ represents the node set in the computation graph of the model. \par
\noindent \textbf{Recursive Case Setup: from p+1 to p}.
For simplicity of notation, we first describe the problem setup of receptive field region calculation in computation graph with only one path including all nodes. The setup can be naturally extended to arbitrary computational graph with the Depth First Search solution. Consider a pure convolutional neural network model with $P$ layer operations and the layer index $p \in[P]$. i.e, for any $p \in[P-1]$, the node representing the $p$th layer in the computation graph has one and only one child node representing the $(p+1)$th layer. The layer operation $type(p)$ can be one of the common operations in Neural Network including \emph{Convolution, Activation, Pooling, Normalization, Interpolation}. Define the feature map $f_{p}$ as the output feature map of the $p$th layer. For any $p \in[P-1]$, We want to derive a recursive and invertible mapping function that maps $r^{p+1}$ with respect to $f_{p+1}$ to $r^{p}$ with respect to $f_{p}$, based on the type of the layer $(p+1)$ and a list of parameters $A_{p+1}$ that characterize the layer $(p+1)$: $r^p = F(r^{p+1},A_{p+1})$

And we formulate detailed recursive cases on convolution, normalization, pooling, and interpolation in Appendix. 

\noindent \textbf{Extended Setup for Arbitrary Computation Graph}.
Suppose the $p$-layer has m children denoted as $p^{l}$ such that $l \in[m]$. 
Denote the receptive field region $r^p$ with respect to the $p$th layer output $f_p$, $r^p = \bigcup_{l \in [m]} F(r^{p^{l}},A_{p^{l}})$

where $F$ denotes the recursive and invertible mapping function that maps receptive field region of child nodes to the receptive field region with respect to current feature map, based on the type of the child layer operation. Note that the union of the outputs of multiple mapping function F might be the union of more than one rectangular regions. In our PyTorch implementation, we approximate the receptive field region $r_p$ by the smallest rectangular region that fully covers all the back-traced rectangular regions before we further compute the receptive field region at a higher nodes. Note that the smallest rectangular region can be trivially computed by simply comparing the coordinate values of sides of all rectangular receptive field regions. The design choice is based on two practical reasons. 1) Multiple back-traced rectangular regions are often greatly overlapped with each other. Thus, negligible theoretical computation overheads are introduced by computing features in the approximated receptive field region; 2) Pytorch implementations are much more efficient in computing features in a rectangular patch. At each layer, exact input and output of individual $F$ function are memorized by the receptive field region controller for AutoCropping in the forwarding pass.

\noindent \textbf{AutoCropping and AutoPadding}.
After (approximate) layer receptive field regions have been computed from leaf to root in the computation graph, we can compute features within the approximate receptive field regions from root to leaf. For any $p$ layer and any child node $p^l$ of the $p$ layer, the receptive field region controller crops the feature map with the memorized output of the $F$ function and pads it with the memorized padding value at four boarders for next-layer feature computation in $p^l$. Any feature value that is outside the approximate layer receptive field $r^p$ does not contribute to final-layer features around the user click. 

\subsection{Conditional ConvNet for RFT}
\noindent \textbf{ConvNet Design}.
\label{sec:conv}
The Conv component of TraceNet includes a feature pyramid backbone that extracts multi-level feature maps from the input image, a conditional filter head that predicts the parameters of the mask head with local features around the click, a compact mask head conditioned on the user-specified instance in their filters. This design is directly inspired from Condinst \cite{condinst}, a dynamic instance-aware ConvNet conditioned on instances.

\noindent \textbf{Backbone Module}.
Following the design of Feature Pyramid Network \cite{FPN}, we extract a 5-level feature pyramid over ResNet \cite{Resnet}. Each level of our feature pyramid is used to extract local features around the click at different scales. The feature pyramid design is essential because no scaling information of the user-specified instance is provided.

\noindent \textbf{Mask Branch Conditioned on Instance}.
The mask branch is in the format of Fully Convolutional Neural Network \cite{FCN} for an image-level prediction. The mask branch is applied to a feature map extracted from the backbone. (i.e  $P_3$ with downsampling ratio of 8). Compared to the mask branch design in Mask RCNN \cite{mask-rcnn}, the design eliminates needs for ROI operations by performing convolution in image-level. Besides, the computation overhead of the mask head in our model is much more lightweight. It only consists of 3 1x1 convolution layers with 8 channels each, while the unconditioned mask branch in Mask-RCNN often has four convolution layers with 256 channels. The intuition behind the compact design can be explained as follows. When the parameters of mask branch in our model are conditioned on local features around user-specified instance, the characteristics (geometry of the instance and relative location of the instance with respect to the user query click) of the user specified instance can be encoded in the mask branch  with the help of conditional filter head. To fully exploit the encoded spatial characteristic, we concatenate the $P_3$ feature map with a map of relative coordinates from all locations to the user click query. Similar designs also exist in other conditional instance segmentation algorithms \cite{condinst, soit}. When applied to the concatenated inputs, the mask branch can naturally focus on the pixels of the user specified instance and predict the mask in an instance-aware manner. A sigmoid is applied make the mask prediction class-agnostic. Finally, a bilinear 4x upsampling is performed on the output mask. The upsampling results in 2x downsampling mask prediction compared to the resolution of the input image. 

\noindent \textbf{Conditional Filter Head and Box Head}.
The conditional filter head is adopted with slight modification from Condinst \cite{condinst}, which is based on FCOS \cite{FCOS}. In FCOS and Condinst, each pixel location on multi-level feature map can be associated with an instance in the original image by a simple mapping. i.e. The pixel $(x,y)$ at the feature map with downsamling ratio of $s$ can be mapped to the imput image as $\left(\left\lfloor\frac{s}{2}\right\rfloor+x s,\left\lfloor\frac{s}{2}\right\rfloor+y s\right)$. Condinst introduces a conditional filter head to encode characteristics of the instance associated with the location in the feature map in pixel-level fashion and a box head to regress the bounding box location of the same instance. The conditional filter head is used to predict a 169-dimension vector of parameters $\boldsymbol{\theta}_{a, b}$ for the above mentioned mask branch for the mask of instance located at $(a, b)$ in the feature map. The box head predicts a 4-dimension vector encoding relative distance between the pixel and four boundaries of the bounding box. Both heads take the features extracted from heavy classification and box tower, which consists of four 3x3 stride 1 convolutional layers with dimension 256 followed by ReLU activation and Batch Normalization. The centerness head is not included in the Figure~\ref{fig:arch} and we refer readers to more details in FCOS \cite{FCOS}.

\section{Experiments}
\vspace{-0.20in}
\begin{table}[ht]
\label{tab:lvis}
\centering
\setlength\tabcolsep{3pt} 
\caption{mIoU-T and mTA on LVIS, mTA on COCO.}
{%
\begin{tabular}{l|lll}
\toprule
Method & mIoU-T & mTA(LVIS) & mTA(COCO)  \\ \midrule
ritm-h32~\cite{RITM}              & \textbf{0.328}  & 0.238 &  0.349\\
focuscut-R-50~\cite{focalcut}              & 0.253  & 0.0734 & 0.117  \\
focuscut-R-101~\cite{focalcut}              & 0.228  & 0.0744 & 0.138  \\
\rowcolor{lightPurple}
TraceNet-R-50-FPN                        & 0.286 & \textbf{0.272} & 0.346\\ 
\rowcolor{lightPurple}
TraceNet-R-101-FPN                        & 0.294 & 0.257 & \textbf{0.395} \\ \bottomrule
\end{tabular}
}
\label{tab:aim}
\end{table}

\begin{table*}[t]
\centering
\caption{mIoU-T over categories on COCO dataset.}
\label{tab:perf_coco}
\vspace{-0.10in}
\begin{tabular}{l|lllllllll}
\hline
\multicolumn{1}{c|}{Method} & \multicolumn{1}{c}{person} & \multicolumn{1}{c}{car} & \multicolumn{1}{c}{chair} & \multicolumn{1}{c}{bottle} & \multicolumn{1}{c}{cup} & \multicolumn{1}{c}{dining table} & \multicolumn{1}{c}{traffic light} & \multicolumn{1}{c}{bowl} & \multicolumn{1}{c}{Category Total} \\
\hline
mobilesam \cite{mobilesam} & 0.4877 & 0.5129 & 0.3530 & 0.6043 & 0.6405 & 0.2571 & 0.3425 & 0.3675 & 0.4331\\
focuscut-R-50 \cite{focalcut} & 0.4128 & 0.3734 & 0.3703 & 0.3705 & 0.4715 & 0.2206 & 0.3652 & 0.4631 & 0.3774\\
focuscut-R-101 \cite{focalcut} & 0.4759 & 0.4150 & 0.3961 & 0.3079 & 0.3610 & 0.2735 & 0.3122 & 0.4411 & 0.4217\\
\rowcolor{lightPurple}
TraceNet-R-50-FPN & 0.4306 & 0.4796 & 0.3285 & 0.090 & 0.1631 & 0.5010 & 0.2925 & 0.2541 & 0.3977\\
\rowcolor{lightPurple}
TraceNet-R-101-FPN & 0.4542 & 0.5186 & 0.3787 & 0.0958 & 0.1798 & 0.5207 & 0.4471 & 0.2733 & 0.4312\\
\hline
\end{tabular}
\end{table*}

\begin{table}[ht]
\label{tab:FLOPS}
\centering
\setlength\tabcolsep{3pt} 
\caption{Computation Cost to Retrieve an Instance from an image of size of 1024 x 768. FullNet refers to our model without RFT.}
\vspace{2mm}
\vspace{-0.20in}
{%
\begin{tabular}{l|lll}
\toprule
Method & throughput (FPS) & latency (ms) & FLOPs \\ \midrule
mobilesam~\cite{mobilesam}        & 40.81  & 24.50 & - \\
ritm-h32~\cite{RITM}              & 11.37  & 87.94 & 406.5G  \\
focuscut-R-50~\cite{focalcut}              & 44.33  & 22.56 & 52.55G  \\
focuscut-R-101~\cite{focalcut}              & 37.86  & 26.42 & 67.18G  \\
FullNet-R-50-FPN                         & - & - & 86.67G \\
FullNet-R-101-FPN                         & - & - & 118.28G \\
\rowcolor{lightPurple}
TraceNet-R-50-FPN                        & \textbf{48.85} & \textbf{20.47} & 34.47G\\ 
\rowcolor{lightPurple}
TraceNet-R-101-FPN                        & 41.96 & 23.83 & 66.67G \\ \bottomrule
\end{tabular}
}
\label{tab:aim}
\end{table}
\vspace{0.15in}
\noindent \textbf{Evaluation Formulation.}
\noindent Denote a set of groundtruth mask $\left\{M_{gt}^{i}\right\}$, where $M_{gt}^{i}$ is the groundtruth mask of the $i$-th instance in the dataset. Denote a set of clicks $\left\{c^{i,j}\right\}$ where each element $c^{i,j}$ is the $j$-th click, $c^{i,j} \in M_{gt}^{i}$. Denote a set of predicted mask as $\left\{M_{pred}^{i,j}\right\}$, where $M_{pred}^{i,j}$ is the predicted mask of the $i$-th instance when being queried by $c^{i,j}$.

\vspace{-0.10in}
\subsection{Evaluation Protocol}\label{sec:protocol}
\vspace{-0.05in}
\noindent Because $M_{pred}$ depends on the user click $c$, we propose a new metric, the mean tap Intersection over Union (mIoU-T), to measure the average segmentation accuracy of all possible user clicks within the groundtruth instance mask. The user click tolerance is measured by a proposed metric mean tap Area (mTA). mTA calculates the ratio between area of feasible clicking area and the area of groundtruth instance mask. The feasible clicking area covers potential clicks that can generate an instance mask with the IoU over a predefined threshold. 

\noindent \textbf{Mean Tap Intersection Over Union}. Since the predicted mask $M_{pred}$ depends on the user click, IoU cannot be directly applied to measure the expectation of mask quality with different user clicks. We propose mIoU-T that measures the average quality of a set of predicted masks $\left\{M_{pred}^{i,j}\right\}$ over a set of instances with groundtruth mask $M_{gt}^{i}$. 
\begin{align}
    mIoU_{T} = \frac{\sum_{i,j}\emph{Area}(M^{i,j}_{\emph{pred}} \cap M^i_{\emph{gt}})}{\sum_{i,j}\emph{Area}(M^{i,j}_{\emph{pred}} \cup M^i_{\emph{gt}})}
\end{align}
\noindent This is a general metric that works on arbitrary numbers of one-shot click queries and arbitrary numbers of instances. In our experiment, the number of clicks is a constant within each instance. i.e for any $i$, $j \in [k]$, where k is a constant value indicating the size of the set of candidate one-shot click.

\noindent \textbf{Mean Tap Area}.
Besides the expectation of predicted mask quality, the user click tolerance is measured, i.e., the proportion of the region that a user click can fall into for a high-quality single-instance mask. We propose to use mTA to measure user click tolerance. To calculate the feasible region of user click, for any $i$, $\left\{c^{i,j}\right\}$ should be constructed as a set of all pixels in the $M_{gt}^{i}$. mTA is calculated as below: 
\begin{align}
    mTA = \frac{\sum_{i,j}\mathbbm{1}(\emph{IoU}(M^{i,j}_{\emph{pred}},M^i_{\emph{gt}}) \geq \beta)}{\sum_{i}\emph{Area}(M^i_{\emph{gt}})}
\end{align}\label{eq:mta} 
\noindent where $\beta$ is a pre-defined IoU threhold and $\mathbbm{1}$ is an indicator function. We use Eq.~\ref{eq:mta} rather than averaging among each of the trials to treat instances of various sizes equally. 
\vspace{-0.10in}
\subsection{Implementation Detail}
\vspace{-0.05in}
\noindent \textbf{Dataset.} We train and evaluate TraceNet on MS-COCO \cite{coco} and LVIS \cite{lvis}.  MS-COCO is a large-scale dataset for object detection and instance segmentation with over 82k training images and 600k instance-level mask annotations. LVIS is a dataset for large vocabulary instance segmentation. It has 2-million mask annotation over 1k entry-level categories.

\begin{figure}[t]
	\centering
	\includegraphics[width=0.5\textwidth]{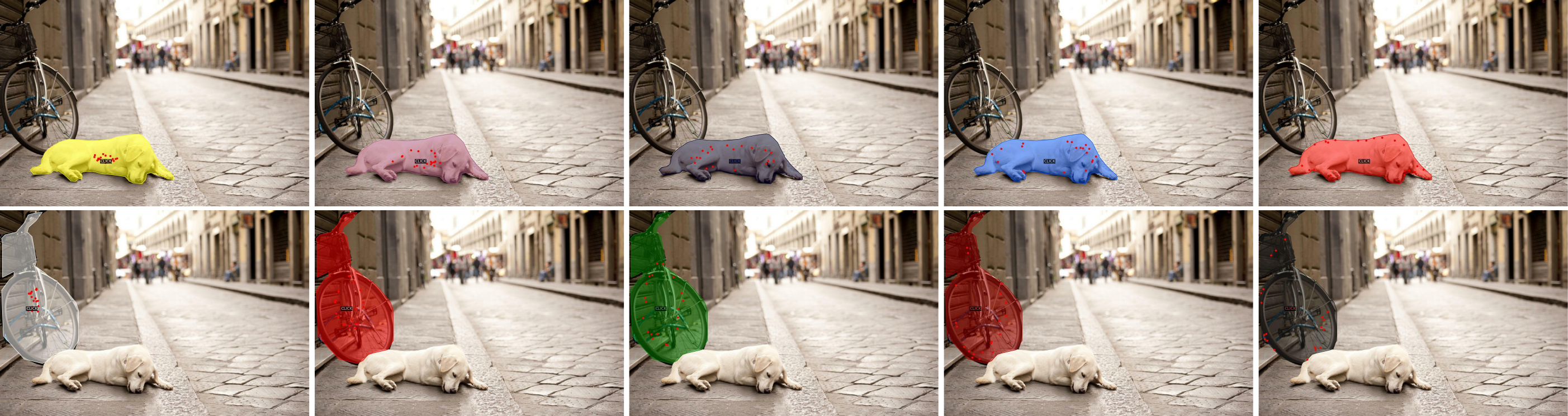}
	\caption{Generated clicks from band-1 (left most) to band-5 (right most).}
	\label{fig:generation}
\end{figure}

\begin{figure}[t]
	\centering
	\includegraphics[width=0.5\textwidth]{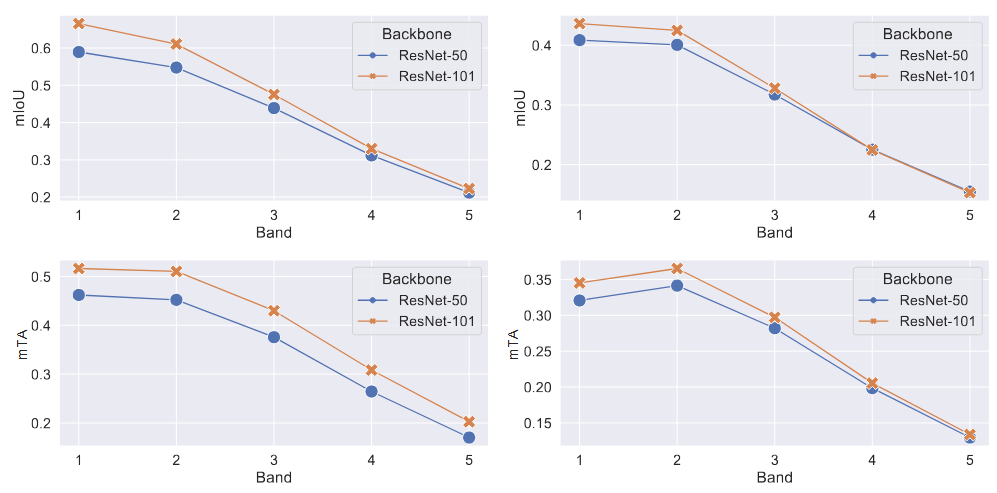}
	\caption{mIoU-T and mTA over different band regions in COCO and LVIS}
	\label{fig:profile}
\end{figure}

\begin{figure}[t]
	\centering
        \vspace{-0.16in}

	\includegraphics[width=0.9\linewidth]{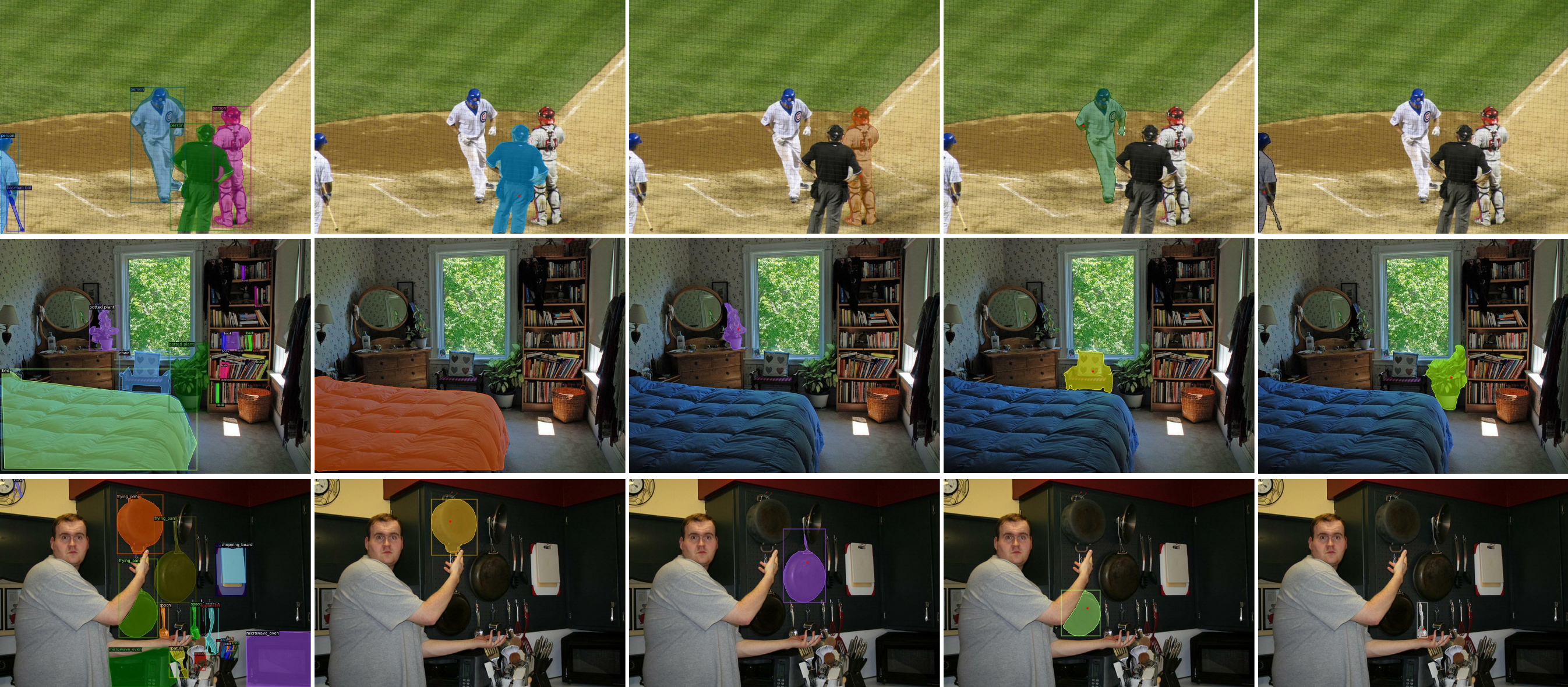}

	\caption{Qualitative Results: first column as groundtruth, others as model predictions. More qualitative results are in Appendix.}
    
	\label{fig:vis}
\end{figure}
\noindent \textbf{Click Simulation with Morphological Transformations}. Since no clicks are provided in the dataset, we have to simulate user clicks within all instances in the dataset. During evaluation, 25 user clicks $\left\{c^{i,j}\right\}$ are simulated  within all instances in the dataset. As is illustrated by the figure in the Appendix, we sample 25 clicks for one instance, with each five clicks randomly sampled in one of the five bands around the moment of the instance. The boundary between bands is constructed with the help of morphological transformations: Suppose the instance is bounded by a bounding box in ${H \times W}$. Then we construct an binary image $I \in\{0,1\}^{H \times W}$, where only the pixel value at moment of the instance is one. The first boundary between bands is generated by performing dilation on the $I$ with kernel size ${H/5 \times W/5}$. More boundaries are generated by performing the same dilation operation repeatedly.  The generated clicks are visualized in Figure \ref{fig:generation}.\par
\noindent \textbf{Training details}. We refer the training protocol, the loss function and hyper-parameters settings to the Appendix, as they are highly similar to CondInst \cite{condinst} and FCOS \cite{FCOS}.

\noindent \textbf{Baselines.} ritm \cite{RITM} is a computationally heavy click based interactive segmentation model with the strongest one click performance before SAM \cite{sam}. focuscut \cite{focalcut} is a state-of-the-art efficient click based interactive segmentation model with the resnet backbone. For fair comparison, focuscut, ritm and our models are trained on LVIS + COCO. Mobilesam is trained on a significant larger dataset proposed in the SAM paper.

\noindent \textbf{mIoU-T.} We evaluate overall IoU-T performance of our method on COCO and LVIS. As shown in the Table 1 and table 2, the TraceNet with ResNet backbone outperforms focuscut with the same backbone. The performance is slightly lower than heavy ritm and mobilesam with more training data. In the category analysis, our model achieves good performance on common and large object categories (e.g dinning table and car) but loses performance in small ones (e.g. bottle and cup) compared to baselines. The intuition is that global features eliminated by RFT module are important when extracting segmentation masks of relatively small objects. 

\noindent \textbf{mTA.} mTAs of TraceNet on COCO are 0.346 and 0.395 with ResNet-50 and RestNet-101 backbone, respectively. It indicates that over 30$\%$ of the user taps can result in a good instance mask with IoU larger than 0.7. A significant drop in terms of mTA is observed on LVIS. Since the LVIS dataset is providing more fine-grained annotations of the image in COCO. Overall, TraceNet achieves significantly better mTA.

\noindent \textbf{Performance Profiling over Band.} The mIoU-T and mTA performance over 5 different bands are profiled in Figure~\ref{fig:profile}. Both mIoU-T and mTA generally drop with increasing distance from tap to the instance center. 

\noindent \textbf{Computation analysis.} Efficiency is one of the most important objectives of one-tap segmentation task for mobile devices. In Table 3, we make an analysis on FLOPs, throughput in one second and latency of inference process of the algorithm. With the help of RFT, TraceNet saves 60.2\% of computations on R-50-FPN Backbone and 43.6\% of computations on R-101-FPN, compared to the full inference process which is denoted as FullNet (for ablation). TraceNet significantly outperforms all baselines in terms of computation efficiency. For memory consumption, the minimum parameters should be stored for inference TraceNet is only 12.5M and is sufficient to meeting the requirements of mobile devices. We include more FLOP results in Appendix with more transformer models.
\vspace{-0.10in}
\section{Conclusion}
\vspace{-0.05in}
In this paper, we propose and formulate click-driven one instance segmentation task as well as design the evaluation protocol. We built a solution TraceNet that back-traces the receptive field region at each layer with respect to local features around the user query tap. Extensive experiments demonstrate effectiveness and efficiency of TraceNet. 

\noindent \textbf{Acknowledgment}. This work was supported by the National Science Foundation grants NSF CNS 21-06592, NSF OAC 18-35834 KN, NSF CNS 19-00875 and NSF CCF 22-17144. Any results and opinions are our own and do not represent views of National Science Foundation.

\bibliographystyle{IEEEbib}
\bibliography{main}

\end{document}